\documentclass{article}
\usepackage[submission]{colm2026_conference}
\usepackage{hyperref}
\usepackage{url}
\usepackage{booktabs}
\usepackage{lineno}
\usepackage{color}
\usepackage{tcolorbox}
\usepackage{caption}
\tcbuselibrary{skins}

\usepackage{xcolor}
\usepackage{graphicx}
\usepackage{tabularx}
\usepackage{amsmath}
\usepackage[capitalize]{cleveref}
\usepackage{arydshln}
\usepackage{mathtools}
\usepackage{multirow}
\usepackage{lmodern}
\usepackage{setspace}
\usepackage{kotex}
\usepackage{array}
\usepackage{newtxtext, newtxmath}
\usepackage{siunitx}
\usepackage{latexsym}
\usepackage[T1]{fontenc}
\usepackage[utf8]{inputenc}
\usepackage{microtype}
\usepackage{inconsolata}
\usepackage{comment}
\usepackage{soul}
\usepackage{makecell}
\usepackage{enumitem}
\usepackage{float}
\usepackage{wrapfig}

\hypersetup{colorlinks,linkcolor={blue},citecolor={blue},urlcolor={blue}}

\newif\ifshowcomments
\showcommentsfalse

\newcommand{\algname}{\texttt{CoRD}}

\definecolor{absgray}{RGB}{242,243,245}
\definecolor{metablue}{RGB}{0,102,204}

\usepackage{pifont}

\newcolumntype{L}[1]{>{\raggedright\let\newline\\\arraybackslash\hspace{0pt}}m{#1}}

\newcolumntype{X}[1]{>{\centering\let\newline\\\arraybackslash\hspace{0pt}}p{#1}}

\usepackage{placeins}

\newcommand{\customabstractpage}{
\begin{tcolorbox}[
    enhanced,
    colback=absgray,
    colframe=absgray,
    boxrule=0pt,
    arc=8pt,
    left=3mm,
    right=3mm,
    top=3mm,
    bottom=3mm
]

{\Large\bfseries
Distilling Long-CoT Reasoning through Collaborative Step-wise Multi-Teacher Decoding
\par}

\vspace{3mm}

Taewon Yun$^{1}$, Jisu Shin$^{1}$, Jeonghwan Choi$^{1}$, Seunghwan Bang$^{2}$, Hwanjun Song$^{1}$\par

\vspace{1mm}

$^{1}$KAIST, $^{2}$UNIST\par

\vspace{4mm}

\noindent
Distilling large reasoning models is essential for making Long-CoT reasoning practical, as full-scale inference remains computationally prohibitive. Existing curation-based approaches select complete reasoning traces post-hoc, overlooking collaboration among heterogeneous teachers and lacking dynamic exploration, which leads to redundant sampling and missed complementary reasoning. We introduce \algname{}, a collaborative multi-teacher decoding framework that performs step-wise reasoning synthesis guided by predictive perplexity–based scoring and beam search. This enables heterogeneous LRMs to jointly construct coherent reasoning trajectories while efficiently preserving diverse, high-potential hypotheses. Experiments show that \algname{} produces higher-quality reasoning data and achieves near teacher-level student performance with fewer, structured supervision signals, without substantial efficiency overhead. \algname{} further generalizes well to out-of-domain and open-ended settings. The dataset and model are available at \href{https://github.com/DISL-Lab/CoRD}{https://github.com/DISL-Lab/CoRD}.
\vspace{4mm}

\noindent
\begin{minipage}[t]{0.65\textwidth}
{\small
\textbf{Date:} April 26, 2026 \par
\textbf{Correspondence:} Hwanjun Song at {\color{metablue}\href{mailto:songhwanjun@kaist.ac.kr}{songhwanjun@kaist.ac.kr}} \par
\textbf{First Author:} Taewon Yun at {\color{metablue}\href{mailto:ytaewon0415@kaist.ac.kr}{ytaewon0415@kaist.ac.kr}} \par
\textbf{Model and Dataset:}{\color{metablue} \url{https://github.com/DISL-Lab/CoRD}}
}
\end{minipage}
\hfill
\begin{minipage}[t]{0.27\textwidth}
\vspace*{-0.1cm}
\raggedleft
\includegraphics[width=1.0\linewidth]{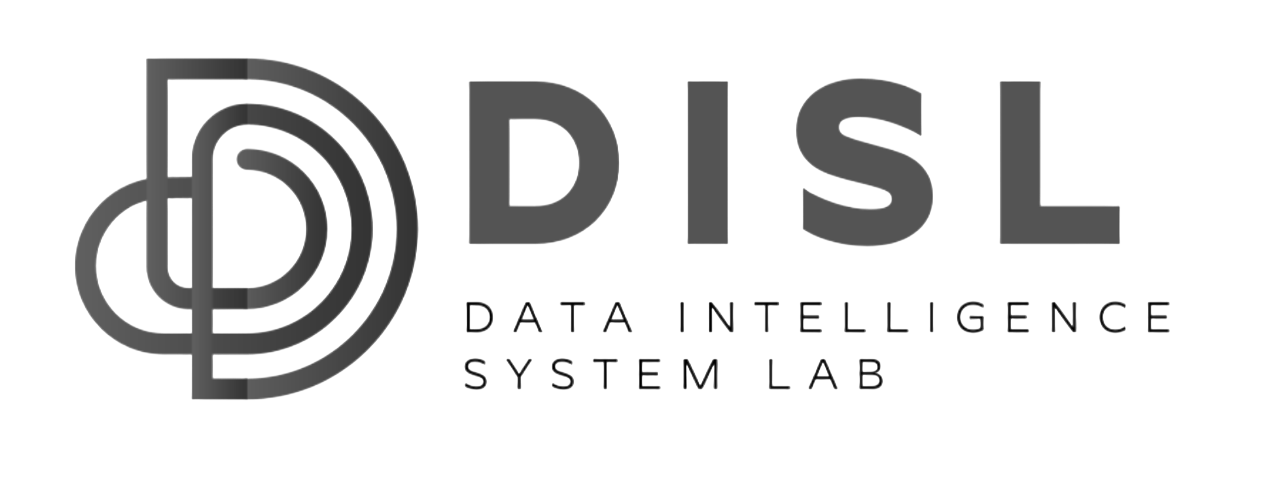}
\end{minipage}

\end{tcolorbox}
}

\begin{document}

\thispagestyle{empty}

\vspace*{-1.3cm}
\customabstractpage

\section{Introduction}
\label{sec:introduction}

% 1) HOMO VS HETERO -> HETERO 시너지 활용 안 한다.
% 2) Curation vs Synthesize -> 현재 Curation 별로다. progress 이상한 거 제거 못한다. 이를 위해 기존 conventional 적용하기에는 LRM 패턴 고려못한다.->

% 2) hetero에서 나오는 collobo 를 잘 활용한다. -> main
% 3) dynamic하게 조절할 수 있다. -> Perplexity
% 4) 
% 1) search 효율적이다

Rapid progress in large reasoning models (LRMs), such as Deepseek-R1 \citep{guo2025deepseek}, has unlocked new capabilities beyond conventional language understanding, enabling complex problem solving \citep{plaat2024reasoning, li2025naturalthoughts}. The key lies in \emph{test-time scaling}, which enhances reasoning by allowing models to deliberate longer, explore broader solution paths, and allocate more computation, often leading to long chain-of-thought (Long-CoT) reasoning \citep{qu2025optimizing, chen2024expanding}.
Yet, the high computational cost and complexity of LRMs hinder deployment, making reasoning distillation into smaller models essential for real-world applications and a growing focus of research \citep{zhang2025s1, li2025naturalthoughts}.

The goal of \emph{reasoning distillation} is to extract high-quality reasoning trajectories from teacher models and transfer them into smaller students through sequence-level knowledge distillation \citep{kim2016sequence}. Here, a reasoning trajectory refers to a structured sequence of intermediate steps, including strategic shifts, reflective self-corrections, and hypothesis revisions, that collectively lead to the final solution. However, identifying high-quality trajectories is particularly challenging in the Long-CoT setting, as they often span thousands of tokens and evolve through dynamic \emph{Aha moments} \citep{guo2025deepseek}. While approaches based on process reward models (PRMs) or Monte Carlo Tree Search (MCTS) are effective for short and static reasoning tasks \citep{park2025ensembling, yao2024mulberry, yin2025towards}, they become impractical when applied to {Long-CoT} reasoning: reward shaping prematurely eliminates reasoning paths that may initially appear suboptimal but are essential for transferring deliberative reasoning patterns, and the search space grows exponentially with trajectory length.

In this regard, recent studies, such as S1 \citep{zhang2025s1} and LIMO \citep{ye2025limo}, have adopted a \emph{curation}-based approach, which first generates complete candidate reasoning traces from multiple (or even identical) teacher models and then selects high-quality ones for distillation. Despite their simplicity, they fail to harness the collaborative potential of multiple heterogeneous teacher models to jointly discover complementary reasoning strategies and compose novel solution paths that no single teacher could produce in isolation. That is, they waste computation on discarded candidates and inherently lack the ability to adjust exploration dynamically due to their post-hoc design.

\begin{figure}[t!]
\begin{center}
\includegraphics[width=10cm]{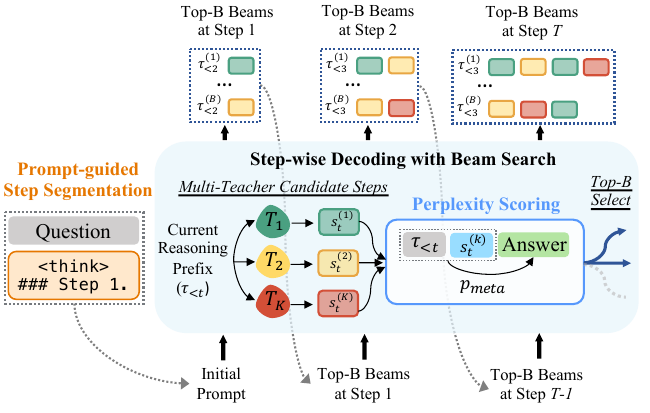}
% \vspace*{-0.5cm} % 하단 여백 추가
\end{center}
\vspace*{-0.45cm}
\caption{Overview of \algname{}: Teacher LRMs collaboratively decode reasoning steps via prompt-guided segmentation. At each step, candidate steps are evaluated via predictive perplexity, retaining the top-$B$ reasoning trajectories for subsequent decoding. The gray dotted line indicates the auto-regressive flow of reasoning steps.} 
\label{fig:overview}
\vspace*{-0.55cm}
\end{figure}

To address these limitations, we propose \algname{} (\underline{Co}llaborative \underline{R}easoning \underline{D}ecoding) in Figure \ref{fig:overview}, a paradigm shift toward a collaborative step-wise decoding process driven by multi-teacher interaction, enabling heterogeneous teachers to jointly construct strategically evolving reasoning trajectories. Instead of generating complete trajectories upfront, \algname{} treats each reasoning step as the minimal unit of generation, allowing teacher LRMs to collaboratively propose and integrate step proposals during their decoding. At each decoding step, we evaluate the quality of the proposed reasoning steps based on a \emph{predictive perplexity score}, which quantifies how well the ground-truth answer is predicted given the current reasoning prefix. This scoring reflects how naturally the reasoning is expected to progress toward the correct solution, enabling early identification and adaptive selection of promising paths without requiring full trajectories. Unlike curation-based approaches, this step-wise evaluation fosters synergistic collaboration among heterogeneous teachers, while unlike MCTS, it avoids repeated rollouts and improves computational efficiency.

Although predictive perplexity offers an effective local signal of short-term consistency, it lacks awareness of long-term payoff. To address this, we integrate \emph{beam search} into our decoding framework, which maintains multiple high-potential trajectories in parallel and preserves reasoning paths that may initially seem sub-optimal but ultimately lead to superior solutions, including strategic shifts and self-corrections often overlooked by reward-driven methods. By formulating reasoning distillation as \emph{step-wise collaborative decoding with beam search}, we transform reasoning from one-shot selection into incremental generation. %, where hypotheses evolve and are refined as decoding unfolds. 
%
%Together, these innovations convert reasoning distillation from a static, offline selection problem into a dynamic, collaborative search process that reduces redundancy, expands exploration, and leverages the complementary strengths of multiple teacher models.

Our evaluation on five close-ended or open-ended reasoning benchmarks compares \algname{} against two multi-teacher distillation baselines (\texttt{Curation} and \texttt{Integration}) as well as state-of-the-art methods (S1 and LIMO-v1/v2), highlighting three main contributions: (1) We propose \algname{}, a novel multi-teacher reasoning distillation framework that reformulates post-hoc reasoning selection into step-wise collaborative decoding, (2) We introduce prompt-guided step segmentation, predictive perplexity scoring, and beam search as core mechanisms, each empirically outperforming alternative designs; and (3) Compared to baselines, \algname{} produces higher-quality reasoning data and distills student models that approach or even surpass teachers.

\section{Related work}
\label{sec:realtedwork}

\noindent\textbf{Test-time Scaling.} 
LLMs achieve stronger performance when their generation is guided by reasoning, such as CoT, rather than directly producing answers \citep{wei2022chain}. Test-time scaling further enhances this ability by allocating additional inference-time computation, enabling models to perform Long-CoT reasoning. 
Techniques like multi-pass inference \citep{he2024enhancing}, which compares multiple attempts, and self-reflection \citep{huang2025efficient, yun2025refeed}, which iteratively revises intermediate steps, have demonstrated substantial improvements. However, these gains entail significant computational overhead \citep{snell2024scaling}, motivating the distillation of test-time scaling capabilities from LRMs into smaller students \citep{yeo2025demystifying}.

% Related work 2
\smallskip\smallskip
\noindent\textbf{Reasoning Distillation.}
Reasoning distillation transfers reasoning from large teacher models to lightweight students by distilling complete reasoning trajectories at the sequence-level, not by token-level logit matching \citep{hu2025distillation, kim2025reinforcement}.
For short-CoT reasoning, PRMs ensure sequence-level quality by filtering incorrect steps \citep{lai2024step, wu2024enhancing}.
MCTS \citep{yao2024mulberry} pairs this correctness-based filtering with exploration, expanding approved steps and synthesizing them into complete reasoning paths.
However, Long-CoT reasoning distillation is more challenging, since PRM overlooks reasoning that could improve through revising intermediate errors, while MCTS struggles with the rapidly expanding search space.
Thus, curation is widely adopted, following a generate-then-select strategy in which LRMs first produce complete reasoning trajectories and then select candidates using simple heuristics \citep{zhang2025s1, ye2025limo}. However, this strategy samples blindly, with no guarantee of valid reasoning or strong training signals, leading to discarded computation.

\smallskip\smallskip
\noindent\textbf{Collaborative Distillation.} 
Distillation from multiple LLMs is a long-standing paradigm, harnessing teacher diversity to curate training data \citep{song2024learning, ma2025communication}.
Beyond diversity, subsequent studies explore collective synergies to produce outcomes unattainable by isolated models. They construct collective responses achieved either through collective MCTS that selects compelling reasoning steps across models \citep{yao2024mulberry} or through direct integration of their responses \citep{wang2024mixture}.
In this vein, recent efforts extend these ideas by leveraging LRMs as additional sources of diversity, but mainly through simple curation \citep{li2025small, ye2025limo}. 
\section{Multi-Teacher Reasoning Distillation}
\label{sec:problem-setup}
%\vspace*{-0.1cm}

We begin by formalizing our multi-teacher reasoning distillation setting. Let \(x\) be a reasoning problem and \(\mathcal{T}\) be a set of \(K\) large reasoning models (LRMs) acting as teachers. In the \emph{curation}-based setting, which has been adopted in prior approaches such as S1 and LIMO, the $k$-th teacher in \(\mathcal{T}\) generates a complete reasoning trajectory \(\tau^{(k)} = (s_1^{(k)}, \dots, s_{T-1}^{(k)}, s_T^{(k)})\) conditioned on the problem \(x\), where \(s_T\) denotes the final answer and the preceding steps ($s_1, \dots, s_{T-1}$) represent a Long-CoT reasoning process\footnotemark[1].
Then, the distillation dataset is constructed by collecting all trajectories generated by the \(K\) teachers and selecting the highest-quality one for each instance based on a quality function \(Q(x, \tau)\). Formally, the final dataset over \(N\) training instances is defined as:
\begin{equation}
\begin{aligned}
&~~~~~~~\mathcal{D}_{\mathrm{curation}} =
\{ 
\big( x_i, \tau(x_i)^{*} \big) 
\}_{i=1}^N, ~~{\rm where}\\
&\!\tau(x_i)^{*} = {\rm argmax}_{\tau^{(k)} \in \{\tau^{(1)}, \dots, \tau^{(K)}\}} Q(x_i, \tau^{(k)}).\!\!
\end{aligned}
\end{equation}
%For instance, S1 selects the trajectory with the highest task accuracy, while LIMO filters trajectories based on reward signals estimated by a preference or process reward model (PRM). 
While this method is simple and effective for selecting high-quality trajectories from multiple teachers, it relies on post-hoc evaluation and thus cannot leverage multiple teacher LRMs to collaboratively explore and refine reasoning paths.

\footnotetext[1]{Each teacher can generate multiple reasoning trajectories, but we assume one per teacher for simplicity of formulation.}

To overcome these limitations, we facilitate \emph{step-wise collaboration} among teacher LRMs during trajectory construction. 
At step \(t\), each teacher proposes a candidate next reasoning step \(s_t^{(k)}\) conditioned on the current prefix \(\tau_{<t}\), 
which consists of all reasoning steps selected before $t$.
Then, a selection criterion \(S(\cdot)\) evaluates each extended trajectory \(\tau_{<t} \oplus s_t^{(k)}\), 
where \(\oplus\) denotes appending a candidate step from any $k$-th teacher to the current reasoning prefix. As a result, the distillation dataset over \(N\) instances is defined as:

{\color{white}placeholder for margin}
\vspace*{-0.7cm}
\begin{equation}
\begin{aligned}
 &~~~\mathcal{D}_{\rm{step-wise}}=
\{ 
\big( x_i, \tau(x_i)^{*} \big) 
\}_{i=1}^N ~~ {\rm where}\\
&~~~~~~~~~~~~~~~ \tau(x_i)^{*} = \Big\{(s_1^{*}, \dots, s_T^{*}) ~|~ \\
&s_t^{*} = {\rm argmax}_{s_t \in \{s_t^{(1)}, \dots, s_t^{(K)}\}} S(\tau_{<t} \oplus s_t^{(k)}),~ \forall t \Big\}.\!\!\!\!\!\!
\end{aligned}
\label{eq:step-wise}
\end{equation}

\vspace*{-0.25cm}
This pipeline enables the composition of complementary reasoning steps from multiple teachers. However, it also introduces challenges such as defining reasoning steps, evaluating their quality, and efficiently managing a larger search space.

\vspace*{-0.05cm}
\section{CoRD: \underline{Co}llaborative \underline{R}easoning \underline{D}ecoding for Reasoning Distillation}
\label{sec:method_framework}
\vspace*{-0.1cm}

To instantiate the step-wise collaboration in Eq.~(\ref{eq:step-wise}), we conceptualize it as a \emph{step-wise auto-regressive decoding} process where each reasoning step acts as a "token" and teacher-proposed steps form the "decoding vocabulary," enabling efficient exploration of a broader search space.

In this section, we present three core components of our approach, \algname{}: (i) Defining consistent steps across diverse Long-CoT trajectories, (ii) Designing a selection criterion to accurately evaluate partial reasoning, and (iii) Capturing global deliberative processes beyond local quality.

%\subsection{CORD's Main Components}
%\label{sec:components}

\vspace*{-0.1cm}
\subsection{Prompt-guided Step Segmentation}
A starting point of step-wise collaborative decoding is addressing the difficulty of consistently segmenting reasoning trajectories into discrete steps, as different LRMs often produce Long-CoT processes with varying granularity, structure, and progression. 
A straightforward solution is the \emph{line-break} step unit \citep{feng2023alphazero, lai2024step}, which segments reasoning at line breaks (\emph{e.g.}, {\textbackslash n\textbackslash n}) into short chunks, offering a uniform structure but little semantic coherence. Similarly, the \emph{prefix}-based approach \citep{li2025think} identifies steps using explicit textual markers (\emph{e.g.}, {wait}), adding semantic cues; however, both the frequency of such markers and the content within each step vary widely across LRMs, hindering direct comparison.

To this end, we introduce \textit{prompt-guided step segmentation}, which inserts explicit markers into the reasoning to divide it into semantically coherent and functionally distinct steps at a consistent level, regardless of the teacher. Specifically, we embed "\texttt{<think> \#\#\# Step}" in the initial prompt, guiding LRMs to structure their reasoning into clearly separated steps during generation. 
This simple yet effective method, shown in Table~\ref{tab:comparision_split}, ensures that each reasoning step is marked and its content logically segmented. As a result, superficial cues such as line breaks or prefix tokens (\emph{e.g.}, {\textbackslash n\textbackslash n} or {wait}) appear naturally within a single step rather than being mistaken for boundaries, enabling more faithful segmentation and reliable cross-model comparison.

\begin{table}[t]
\centering
\renewcommand{\arraystretch}{1.3}
\footnotesize
\begin{tabular}{|p{13.5cm}|}
\specialrule{\heavyrulewidth}{0pt}{0pt}
Prompt-guided LRM Reasoning \\
\specialrule{\heavyrulewidth}{0pt}{0pt}
\colorbox{cyan!30}{\#\#\# Step 1. Understanding the problem} 
Okay, so we have four circles that are all mutually externally tangent~$\cdots$ \colorbox{red!30}{\textbackslash n\textbackslash n} 
\colorbox{cyan!30}{\#\#\# Step 2. Recalling Descartes' Circle Theorem} 
Descartes' Circle Theorem relates the curvatures of four mutually tangent circles. The formula is:
\colorbox{red!30}{\textbackslash n\textbackslash n} 
$k_4 = k_1 + k_2 + k_3 \pm 2\sqrt{k_1k_2 + k_2k_3 + k_3k_1}$
\colorbox{red!30}{\textbackslash n\textbackslash n} 
Where k = 1/r for each circle~$\cdots$
Hmm, but \colorbox{yellow!30}{wait}, if you have four circles all externally tangent, one of them might be the outer circle enclosing the other three. \colorbox{yellow!30}{Wait}, maybe~$\cdots$ \\
\specialrule{\heavyrulewidth}{0pt}{0pt}
\end{tabular}
\vspace{-0.3cm}
\caption{Comparison of step segmentation. 
\colorbox{red!30}{Red}, \colorbox{yellow!30}{Yellow}, and \colorbox{cyan!30}{Blue} 
represent line-break, prefix, and prompt-guided segmentation, respectively.
}
\vspace{-0.4cm}
\label{tab:comparision_split}
\end{table}

\subsection{Perplexity-based Step Selection}
\label{sec:perplexity}
Another crucial aspect is defining the selection criterion 
$S(\cdot)$ in Eq.~(\ref{eq:step-wise}), which decides the most promising candidate step among those proposed by teacher LRMs.
Thus, we view collaborative decoding as a \emph{step-level} extension of the autoregressive decoding framework: At each decoding step \( t \), each teacher generates a candidate reasoning step \( s_t^{(k)} \) conditioned on the current prefix \( \tau_{<t} \). These $K$ proposals collectively form the \emph{decoding vocabulary}  
$\mathcal{V}_t = \{ s_t^{(1)}, s_t^{(2)}, \dots, s_t^{(K)} \}$,
where the conventional notion of a token vocabulary is replaced by a set of reasoning steps proposed by multiple teachers.

For the scoring function, we introduce a separate model, referred to as the \emph{meta-prover} (MP), which estimates the conditional probability of the ground-truth answer given a partial reasoning trajectory (See Appendix \ref{appendix:data_generation_param} for the prompt used to compute perplexity)\footnotemark[2]. Specifically, at decoding step \( t \), let \( \tau_{<t} \) denote the reasoning prefix up to the previous step, and \( s_t^k \) a next reasoning step proposed by the $k$-th teacher LRM. When this step is appended, the updated reasoning state becomes \( \tau_{<t} \oplus s_t^k \). Then, the meta-prover $p_{meta}$ models the joint conditional probability of the answer tokens given this updated prefix, from which the \emph{predictive perplexity score} used to evaluate candidate steps is derived as:
\begin{equation}
\!\!S(\tau_{<t} \oplus s_t^{(k)})\!=\!{\rm exp} \Big( \frac{1}{M} {\rm log}~ p_{\text{meta}}(A\!\mid\!\tau_{<t} \oplus s_t^k )\Big)
\label{eq:selection}
\end{equation}
\begin{equation*}
\!\!p_{\text{meta}}(A\!\mid\!\tau_{<t}\!\oplus\!s_t^k )\!=\! \prod_{m=1}^M \!p_{\text{meta}}\big(a_m\!\mid\!\tau_{<t}\oplus s_t^k, a_{<m}\big)\!
\end{equation*}
where \( A = (a_1, \dots, a_M) \) denotes the ground-truth answer represented as a sequence of tokens, {yielding a bounded predictive perplexity score in the \([0, 1]\)}.

\footnotetext[2]{We use QwQ-32B as the meta-prover, the strongest LRM in our pool: \{R1-Qwen-32B, QwQ-32B, Phi4-Reasoning-Plus\}; results with other meta-provers are reported in Appendix~\ref{appendix:other_meta_prover}.}

That is, the selected step is determined by the predictive perplexity score, where a higher value indicates that the extended reasoning trajectory better predicts the correct answer. Thus, the step with the highest score $s_t^*$ is chosen from the entire decoding vocabulary $\mathcal{V}_t$ at time $t$.

%
% {\color{blue}
% Its effects are compared against other selection strategies in Section~\ref{sec:component_wise}.
% }

\subsection{Step-wise Decoding with Beam Search}
The selection in Eq.~(\ref{eq:selection}) unfolds auto-regressively, progressively extending the reasoning trajectory until the special \texttt{</think>} token signals its completion. When the pre-defined token budget is exhausted before this point, the sequence is terminated by appending \texttt{</think>} immediately. The teacher selected at the final decoding step generates the final answer based on the completed reasoning.

However, the greedy decoding above suffers from a fundamental limitation. By always choosing the locally optimal step, it can prematurely commit to sub-optimal paths, discarding alternatives that enable strategic shifts to emerge later in Long-CoT reasoning. 
{
On the other hand, MCTS estimates global utility by rolling out complete reasoning trajectories at each step, it becomes computationally prohibitive for Long-CoT reasoning due to the extensive search space. 
}
To address the trade-off between them, we integrate \emph{beam search} into our decoding pipeline, which maintains the top-\(B\) most promising partial reasoning trajectories at each step instead of pursuing a single path.  
At decoding step \(t\), we denote the beam from the previous step as \(\mathcal{B}_{t-1} = \{ \tau_{<t}^{(1)}, \tau_{<t}^{(2)}, \dots, \tau_{<t}^{(B)} \}\).  
The beam is then updated by extending every prefix with candidate steps from its decoding vocabulary \(\mathcal{V}_t^{(b)}\), producing a total of $B\times K$ proposals at step $t$. From these, the top-\(B\) updated trajectories with the highest predictive perplexity scores are selected:
\begin{equation}
\mathcal{B}_t = \text{Top-}B \big( \mathcal{C}_t \big) ~~{\rm where} 
\end{equation}
\begin{equation*}
\mathcal{C}_t = \{\tau_{<t}^{(b)} \oplus s_t^{(k)} \mid \tau_{<t}^{(b)}\in\mathcal{B}_{t-1}, ~s_t^{(k)} \in \mathcal{V}_{t}^{(b)} \}.
\end{equation*}
Compared to greedy decoding, beam search retains alternative reasoning paths that enable strategic shifts, with more reasonable overhead than MCTS.

\subsection{Computational Complexity}
\label{sec:computation}

We analyze the computational complexity of \algname{} using Big-O notation and compare it with greedy decoding and MCTS, as well as the curation-based method, to clarify the computational overhead by its step-wise generation and beam search.

%\algname{} aims to improve reasoning quality while incurring additional compute  due to beam search. To assess this overhead, we analyze its Big-O complexity and compare it to greedy decoding and MCTS. % as well as curation. 
%  and, under a matched compute budget, 

Let $T$ denote the length of a reasoning trajectory (\emph{i.e.}, the number of generated steps), and let $M$ denote the meta-prover cost. For a fair comparison consistent with our experimental setup, all methods generate $B$ reasoning trajectories in total.

\smallskip
\noindent\textbf{CoRD.} At each decoding step, \algname{} generates a total of $K \times B$ proposals and scores them using the meta-prover. With cached key-value states, each expansion requires only an incremental forward pass, yielding $KMB$ expansions per step and an overall complexity of $\mathcal{O}(T K M B)$. The greedy decoding is a special case of \algname{} with beam size $=1$. %, and thus can be repeated up to $B$ times under the same budget. 

\smallskip
\noindent\textbf{MCTS.} This retains a single reasoning trajectory at each step, but it estimates rewards via full rollouts every step. As rollouts complete the remaining trajectory from $\tau_{<t}$, their expected cost decreases with depth and is approximated as $\log (T)$. Repeating this process up to $B$ runs under the budget leads to a total complexity of $\mathcal{O}\big(T K \log (T M B ) \big)$. 

\smallskip
\noindent\textbf{Curation.} Curation generates full reasoning trajectories in a single pass. Each of the $K$ teachers produces a trajectory of length $T$, scored once by the meta-prover after generation. This can be repeated up to $B$ rollouts under the budget, after which the highest-scoring trajectory is selected post-hoc. This results in a total complexity of $\mathcal{O}(TKB)$
%, while the $M$ overhead is negligible.

Taken together, \algname{} incurs lower complexity than MCTS. Although it is more expensive than greedy decoding or curation, we show that (i) \algname{} yields higher-quality Long-CoT trajectories that cannot be obtained by simply increasing the sample budget of greedy decoding or curation, enabled by step-wise collaborative decoding, (ii) the meta-prover overhead ($M$) is negligible in practice. See details in Section \ref{sec:additional_study_search} for reasoning quality and Appendix \ref{appendix:computational_efficiency} for efficiency under an identical answer-reaching sample budget, respectively.

\section{Evaluation}
\label{sec:experiment}

% --- Section 4 보면서 수정하기!
% RQ1. 데이터 퀄리티 관점에서 robust하게 좋은 데이터를 생성해낼 수 있는지. (search 가 잘 성공할 수 있는지)

% RQ2. distillation performance 관점에서 제일 heterogeous LRM의 이점을 잘 살린게 어떤건지, homogenuous 로는 달성할 수 없는.. 그리고 curated dataset에 상관없이 항상 뭐 디스틸레이션 퍼포먼스를 잘 유지하는지?

In this section, we evaluate the quality of reasoning data generated by \algname{} and the performance of a student model trained on it, demonstrating how reasoning quality influences final task outcomes.

\smallskip
\noindent\textbf{Baselines.} We compare \algname{} against two baselines, \texttt{Curation} and \texttt{Integration}, both of which leverage multiple teachers for reasoning distillation using a post-hoc approach, as detailed below.

$\bullet$~ {\texttt{Curation}}: This pipeline is the standard approach used in S1 and LIMO, where each teacher LRM generates a complete trajectory, all are scored as a whole, and the highest-scoring one is selected. For fairness, we apply the same scoring in Eq.~(\ref{eq:selection}).

% , where each teacher LRM generates a complete trajectory, all are scored as a whole, and the highest-scoring one is selected.

% , where each teacher LRM generates a complete trajectory, all are scored as a whole, and the highest-scoring one is selected.

$\bullet$~ {\texttt{Integration}}: This pipeline performs a post-hoc process in which an external integrator (GPT5-mini) merges the complete reasoning trajectories generated by multiple teachers into a single trajectory, selecting and combining consistent parts from each. Refer to the merging prompt in Table \ref{tab:integrated-reasoning-prompt}.

The key distinction of these methods, including \algname{}, lies in whether multiple teachers are used independently (\texttt{Curation}), merged post-hoc (\texttt{Integration}), or collaboratively combined during step-wise collaborative decoding (\algname{}).

\smallskip
\noindent\textbf{Teacher Configuration.} We consider two multi-teacher configurations: \emph{(i) homogeneous}, where all teachers share the same architecture but differ due to sampling with different temperatures in \{0.5, 0.6, 0.7\}; and \emph{(ii) heterogeneous}, where teachers vary in architecture to provide complementary reasoning. For the homogeneous setup, we fix the teacher LRM as QwQ-32B, while for the heterogeneous setup, we additionally include R1-Distil-Qwen-32B (abbreviated as R1-Qwen-32B) and Phi4-Reasoning-Plus alongside QwQ-32B. The sampling temperature is fixed at 0.6 for the three teachers. 

\smallskip
\noindent\textbf{Reasoning Data Distillation.} To distill Long-CoT reasoning, we use the LIMO-v1 dataset, which contains $817$ question–solution pairs curated from millions of mathematical problems via multi-stage filtering based on difficulty and reasoning depth. We then augment the reasoning traces over the dataset using two baseline pipelines\footnotemark[3], including \algname{}, and train three student models, R1-Qwen-7B/14B/32B, through supervised fine-tuning on each of the constructed datasets. All the trained students are evaluated on two widely used mathematical reasoning benchmarks, AIME24 and AIME25. Refer to Appendix \ref{appendix:train_and_inference_param} for detailed training configurations.

%Furthermore, we compare the quality of the data constructed by \algname{} with that of S1 and LIMO-v1/v2 in Section \ref{sec:comp_exist}.

\footnotetext[3]{For reasoning generation, we set the maximum output to $20{,}480$ tokens, allocating $16{,}384$ for \texttt{<think>} reasoning and $4{,}096$ for the final answer to prevent overthinking.}

\smallskip
\noindent\textbf{Hyperparameters.} We set the beam size of \algname{} to 4, producing four partial trajectories at each decoding step. For a fair comparison under the same compute budget, we equalize the total number of generated reasoning trajectories across \texttt{Curation} and \texttt{Integration} by adjusting the number of rollouts, generating four complete trajectories per teacher.

%---

%We design experiments to examine how different multi-LRM frameworks organize reasoning collaboration under an identical compute budget and how such organization affects reasoning quality and distillation performance.  
%
%We compare three reasoning distillation framework—Curation, Integration, and \algname{}—that differ in their dependency and collaboration dynamics, \textit{i.e.}, whether LRMs reasoning generate independently or dependently, and whether the collaboration occurs post-hoc or during decoding.
%
%We further investigate which framework can realize the collaborative synergy that homogeneous configurations fail to achieve.

%
% Predictive perplexity-based scores are evaluated using the same model type in homogeneous settings, and with strongest models in heterogeneous settings.  
%

%

\subsection{Reasoning Quality Comparison}

We evaluate the quality of the generated reasoning across three pipelines: \texttt{Curation}, \texttt{Integration}, and \algname{}. A high-quality Long-CoT reasoning is expected to satisfy two criteria: \emph{(i) answer accuracy}, where {the final answer in the reasoning trajectory matches the ground-truth}, ensuring task correctness, and \emph{(ii) predictive perplexity}, where the predictive perplexity conditioned on the reasoning is high, reflecting progress consistency. Table~\ref{table:method-comparison-final} compares Long-CoT reasoning quality across three distillation pipelines under homogeneous and heterogeneous teachers. While all teachers are fixed to QwQ-32B in the homogeneous setup, additional results with alternative teacher choices are in Appendix \ref{appendix:results1_remaing_results}.

%\vspace*{-0.05cm}
\smallskip\smallskip\noindent
{{\textbf{Highlight.}}} \algname{} achieves the highest answer accuracy and predictive perplexity for its generated Long-CoT reasoning, with the advantage becoming more pronounced under the heterogeneous setup, where diverse teacher signals with complementary reasoning styles interact step by step to reinforce each other, suppress unstable trajectories early, and explore alternative solution paths. 
This leads to a richer and more consistent reasoning dynamics than in the homogeneous setting, where teachers offer limited diversity despite temperature variations.

\begin{wraptable}{r!}{0.5\textwidth} 
\vspace*{-0.3cm}
\renewcommand{\arraystretch}{1.1}
\footnotesize
\begin{tabularx}{0.5\textwidth}{|c |l|c|c|}
\specialrule{\heavyrulewidth}{0pt}{0pt}
Teacher & Distillation  & Answer & Predictive \\ 
Config. & Pipeline  & Accuracy & Perplexity \\ 
\specialrule{\heavyrulewidth}{0pt}{0pt}
\multirow{3}{*}{Homo.} 
  & \texttt{Curation}    & 77.4  & 0.664 \\ 
  & \texttt{Integration} & 88.6  & 0.215 \\ 
  & \textbf{\algname{}}  & \textbf{90.0}  & \textbf{0.726} \\ 
\specialrule{\lightrulewidth}{0pt}{0pt}
\multirow{3}{*}{Hetero.} 
  & \texttt{Curation}    & 84.8  & 0.652 \\ 
  & \texttt{Integration} & 91.2  & 0.223 \\ 
  & \textbf{\algname{}}  & \textbf{93.1}  & \textbf{0.774} \\ 
\specialrule{\lightrulewidth}{0pt}{0pt}
\end{tabularx}
\vspace*{-0.2cm}
\caption{Quality of the generated reasoning across three distillation pipelines under two teacher configurations: Homogeneous (Homo.) and Heterogeneous (Hetero.). Best values for each setup are highlighted in bold.}
\label{table:method-comparison-final}
\vspace*{-0.4cm}
\end{wraptable}

%\vspace*{-0.05cm}
\smallskip\smallskip\noindent
{{\textbf{Detailed Analysis.}}} The observed quality gap can be attributed to two fundamental aspects of multi-teacher distillation: \emph{(i) complementarity exploitation}, which concerns how effectively diverse reasoning signals are combined, and \emph{(ii) collaborative composition}, which captures how those signals interact during the reasoning process itself.

First, regarding {complementarity exploitation}, the strength of \algname{} is evident when contrasted with \texttt{Curation}. The latter follows a generate-then-select strategy, where each teacher produces complete reasoning independently and complementary signals are never exchanged, leading to the lowest answer accuracy. In contrast, \algname{} integrates signals via step-wise collaborative decoding, improving both metrics by reinforcing complementary reasoning
% and filtering unstable paths early.

Second, in terms of {collaborative composition}, \algname{} differs fundamentally from \texttt{Integration}. While this post-hoc fusion improves answer accuracy over \texttt{Curation}, it cannot shape the reasoning process and rather compresses it into less deliberative Short-CoT forms, leading to very low predictive perplexity. In contrast, \algname{} composes reasoning incrementally, allowing complementary signals to guide each step and yielding deeper, more coherent trajectories preserving benefits of test-time scaling.

\begin{wrapfigure}{r}{0.5\textwidth}
\vspace*{-0.3cm}
\begin{center}
\includegraphics[width=0.53\textwidth]{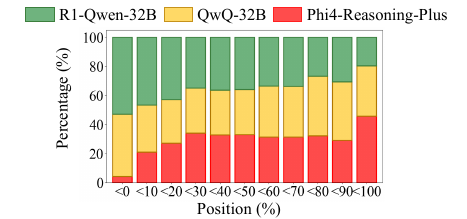}
\end{center}
\vspace*{-0.3cm}
\caption{Teacher selection hit rates (\%) in \algname{} over reasoning progress where decoding steps are mapped to a 0–100\% scale to align varying trajectory lengths.}
\label{fig:hit-rate-solo}
\vspace*{-0.3cm}
\end{wrapfigure}

\smallskip\smallskip\noindent 
\textbf{Analysis of Collaboration Dynamics.} To understand the source of \algname{}'s advantage, we examine teacher selection hit rates in Figure~\ref{fig:hit-rate-solo}, which measure how often each teacher's step candidate is selected over normalized reasoning progress. \algname{} exhibits a specialized allocation pattern, where each teacher is selected for the reasoning phase that best matches its strengths rather than being uniformly sampled from a shared pool. R1-Qwen-32B and QwQ-32B dominate selection in the early phases ($\leq$40\%), which correspond to problem formulation and constraint analysis, while Phi4-Reasoning-Plus increasingly takes over in the late phases ($\geq$80\%), where prior steps must be synthesized into a conclusion. Such specialization is possible because each teacher conditions on the shared prefix $\tau_{<t}$ accumulated from prior steps, so the predictive perplexity scoring captures not only local step quality but also how well a candidate aligns with the current trajectory context. Prompt-guided step segmentation and beam search further reinforce these collaboration dynamics, producing more distinct and stable specialization patterns compared to their counterparts (see Appendices~\ref{app:step_unit_example} and \ref{appendix:decoding_strategy}).

\begin{table}[t]
\vspace{4pt}
\centering
\footnotesize
\setlength{\tabcolsep}{7pt}
\renewcommand{\arraystretch}{1.15}

\begin{tabular}{|l|ccc|ccc|}
\Xhline{1pt}
\multicolumn{7}{|c|}{Teacher Model Performance} \\
\hline
Model Name & \multicolumn{3}{c|}{AIME24} & \multicolumn{3}{c|}{AIME25} \\
\hline
R1-Qwen-32B         & \multicolumn{3}{c|}{71.6} & \multicolumn{3}{c|}{53.8} \\
QwQ-32B             & \multicolumn{3}{c|}{77.9} & \multicolumn{3}{c|}{66.7} \\
Phi4-Reasoning-Plus & \multicolumn{3}{c|}{78.9} & \multicolumn{3}{c|}{67.9} \\
\hline \hline

\multicolumn{7}{|c|}{Student Model Performance (R1-Qwen 7B / 14B / 32B)} \\
\hline
\multirow{2}{*}{Distillation Pipeline} 
 & \multicolumn{3}{c|}{AIME24} 
 & \multicolumn{3}{c|}{AIME25} \\
\cline{2-7}
 & 7B & 14B & 32B & 7B & 14B & 32B \\
\hline

w/o Distillation           & 51.3 & 68.1 & 71.6 & 37.5 & 50.6 & 53.8 \\
\hline

\texttt{Curation-Homo}     & 55.8 & 72.5 & 74.2 & 40.2 & 54.7 & 62.7 \\
\texttt{Integration-Homo}  &  7.9 &  7.1 & 11.9 &  5.4 &  6.3 &  6.9 \\
\textbf{\algname{}-Homo}   & \textbf{58.5} & \textbf{73.7} & \textbf{75.8}
                           & \textbf{42.9} & \textbf{59.3} & \textbf{64.4} \\
\hline

\texttt{Curation-Hetero}   & 56.6 & 68.1 & 75.0 & 42.1 & 54.6 & 62.1 \\
\texttt{Integration-Hetero}&  8.3 &  7.5 & 12.7 &  3.8 &  4.0 &  9.0 \\
\textbf{\algname{}-Hetero} & \textbf{60.8} & \textbf{74.8} & \textbf{79.6}
                           & \textbf{45.6} & \textbf{62.3} & \textbf{70.2} \\
\Xhline{1pt}
\end{tabular}
\vspace{0.1cm}
\caption{Distillation performance comparison across three pipelines under two teacher configurations. The upper block reports teacher performance, while the lower block shows student performance on AIME24 and AIME25 with and without reasoning distillation.}
\label{table:model-size-comparison-final}
\vspace{-0.3cm}
\end{table}

We evaluate how the reasoning quality in Table \ref{table:method-comparison-final} translates into student model performance after distillation. Following recent protocols \citep{ye2025limo, guo2025deepseek}, the student’s reasoning and answers are generated with a maximum output length of $32{,}784$ tokens and a temperature of 0.6. We report Pass@1, the proportion of test questions where the model’s first generated answer matches the correct solution, on AIME24 and AIME25. Pass@1 is computed as the average accuracy over 16 runs to ensure a stable performance estimate. Table \ref{table:model-size-comparison-final} shows the Pass@1 of student models (R1-Qwen-7B/14B/32B) with different sizes, each trained using three distillation pipelines under two teacher configurations. 

\smallskip\smallskip\noindent
{{\textbf{Result.}}} \algname{} consistently delivers the highest Pass@1 across all student model sizes and teacher settings, demonstrating substantial improvements over the two baselines. The gains are particularly pronounced under heterogeneous teachers, where \algname{} effectively integrates complementary reasoning signals. Remarkably, the 32B student model distilled with \algname{} even surpasses the performance of all individual teacher models on both AIME24 and AIME25, indicating that the collaborative signals distilled through step-wise reasoning go beyond simple teacher imitation. This highlights the ability of \algname{} to preserve and enhance high-quality reasoning patterns during training, enabling students to approach or exceed teacher-level performance. 

\smallskip\smallskip\noindent
{{\textbf{Relation to Quality Metrics.}}} 
The performance trends align closely with the reasoning quality analysis in Table \ref{table:method-comparison-final}. Predictive perplexity strongly correlates with student performance, as it captures how well the reasoning guides the model toward the correct solution. In contrast, answer accuracy, which focuses only on the final outcome, fails to translate into comparable gains, as seen in \texttt{Integration}, which achieves higher accuracy than \texttt{Curation} but yields significantly poorer distillation performance because it collapses reasoning into short-form CoT and loses valuable supervision signals. This trend is consistent across another student architecture and a stronger integrator (see Appendices~\ref{appendix:results1_other_model}--\ref{appendix:result_strong_integration}).

\begin{wrapfigure}{r}{0.55\textwidth}
\vspace*{-1cm}
\begin{center}
\includegraphics[width=7.7cm]{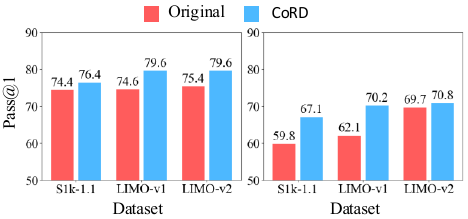}
% \vspace*{-0.5cm} % 하단 여백 추가
\end{center}
\vspace*{-0.2cm}
{\small \hspace*{1.4cm} (a) AIME24. \hspace*{1.7cm} (b) AIME25.}
\vspace*{-0.3cm}
\caption{Performance comparison of student models trained on \algname{}’s reasoning data and curated datasets from S1k-1.1 and LIMO-v1/v2, respectively.}
\label{fig:dataset-student-comparison}
\vspace*{-0.3cm}
\end{wrapfigure}

\smallskip\smallskip\noindent
{{\textbf{Comparison with S1 and LIMO.}}} 
We compare our reasoning data with prior curation-based datasets, S1k-1.1 and LIMO-v1/v2. This comparison highlights the advantage of our collaborative decoding over static curation, showing its ability to generate higher-quality reasoning that yields stronger, more stable distillation. By applying \algname{} to base datasets with varying sizes and question distributions (S1k-1.1 with $1{,}000$ questions, LIMO-v1 with 817, and LIMO-v2 with 800), we demonstrate consistent performance gains regardless of the dataset. Figure \ref{fig:dataset-student-comparison} presents the Pass@1 comparison, where the same student model (R1-Qwen-32B) is trained on equal amounts of data from either the original curated reasoning or our \algname{}, ensuring a fair comparison.

The student model distilled on \algname{} outperforms those trained on the original datasets on both benchmarks, with particularly larger gains on AIME25, which is more challenging. These results show that while curation-based approaches rely on manual dataset design and filtering, step-wise decoding like \algname{} automatically produces higher-quality reasoning data and improve distillation performance.

\subsection{Component-wise Analysis}
\label{sec:component_analysis}

\algname{} has three key components for effective and efficient Long-CoT reasoning synthesis: (i) prompt-guided step segmentation, (ii) perplexity-based step selection, and (iii) decoding with beam search. We evaluate each component individually to understand its contribution to the overall performance.

\begin{wraptable}{r!}{0.55\textwidth}
\vspace*{-0.5cm}
\renewcommand{\arraystretch}{1.1}
\footnotesize
\begin{tabular}{|L{1.9cm}|X{0.8cm}|X{0.8cm}|X{1.0cm}|X{1.0cm}|}
\specialrule{\heavyrulewidth}{0pt}{0pt}
{Segmentation} &
\multicolumn{2}{c|}{Reasoning Qual.} &
\multicolumn{2}{c|}{Distillation Perf.} \\
Method & Acc.
& PP.
& \makecell{\!\!AIME24\!\!}
& \makecell{\!\!AIME25\!\!} \\
\specialrule{\heavyrulewidth}{0pt}{0pt}
Line-break           & 88.4  & 0.734  & 76.7 & 67.7 \\
Prefix               & 91.3  & 0.747  & 77.1 & 67.3 \\
\textbf{Prompt-guide} & \textbf{93.1} & \textbf{0.774} & \textbf{79.6} & \textbf{70.2} \\
\specialrule{\lightrulewidth}{0pt}{0pt}
\end{tabular}
\vspace*{-0.2cm}
\caption{Comparison of \algname{} across three step units (Acc.=answer accuracy; PP.=predictive perplexity).}
\label{table:split-comparison}
\vspace*{-0.3cm}
\end{wraptable}

\subsubsection{Effect of Step Segmentation} 
We examine how different step units in \algname{}’s collaborative decoding affect reasoning quality and distillation performance under the heterogeneous teacher setup. We compare our prompt-guided step segmentation with two alternatives, \emph{line-break} and \emph{prefix-based} methods (see Appendix \ref{app:step_unit_example} for details). Table \ref{table:split-comparison} compares three step-segmentation variants for the student model R1-Qwen-32B. The prompt-guided step unit proves most effective, capturing both style consistency and semantic parity that enable multiple teacher LRMs to reason within a shared step. In contrast, the prefix-based approach aligns better with semantic boundaries but lacks style consistency, while the line-break approach maintains style consistency but fails to achieve semantic alignment, limiting collaborative synergy. These results demonstrate that well-structured step segmentation is essential for maximizing multi-teacher collaboration and producing high-quality supervision signals.

\subsubsection{Effect of Step Selection Criterion}
To understand the impact of selection strategies in \algname{}, we compare our predictive perplexity-based selection with four alternatives: two trajectory-level approaches that select the entire reasoning post-hoc (Random Selection and Max-length Selection) and two step-wise approaches that use either a Process Reward Model (PRM) based on Qwen2.5-Math-PRM-72B or Binary Judgment from LLMs.

\begin{wraptable}{r!}{0.55\textwidth}
\vspace*{-0.3cm}
\renewcommand{\arraystretch}{1.1}
\footnotesize
\begin{tabular}{|L{2.5cm}|X{0.7cm}|X{0.6cm}|X{0.8cm}|X{0.8cm}|}
\specialrule{\heavyrulewidth}{0pt}{0pt}
{\!\!Selection} &
\multicolumn{2}{c|}{\!\!Reasoning Qual.\!\!} &
\multicolumn{2}{c|}{\!\!Distillation Perf.\!\!} \\ 
\!\!Method &
Acc. &
PP.\!\!\!\!&
\makecell{\!\!\!\!AIME24\!\!\!\!} &
\makecell{\!\!\!\!AIME25\!\!\!\!} \\ 
\specialrule{\heavyrulewidth}{0pt}{0pt}

% \multicolumn{5}{|c|}{Trajectory-level Selection} \\ 
% \midrule
\!\!Random Selection      & \!80.4\! & 0.494\!\!\!\! & \!69.0\! & \!61.9\! \\ 
\!\!Max-length Selection\!\!  & \!80.0\! & 0.502\!\!\!\! & \!68.8\! & \!59.0\! \\ 

% \midrule
% \multicolumn{5}{|c|}{Step-level Selection} \\ 
% \midrule
\!\!PRMs                  & \!82.6\! & 0.591\!\!\!\! & \!75.0\! & \!64.6\! \\ 
\!\!Binary Judgment      & \!91.7\! & 0.626\!\!\!\! & \!77.7\! & \!66.3\! \\ 
\textbf{\!\!Predictive Perplexity\!\!\!} 
                      & \!\textbf{93.1}\! & \textbf{0.774}\!\!\!\! 
                      & \!\textbf{79.6}\! & \!\textbf{70.2}\! \\ 
\specialrule{\heavyrulewidth}{0pt}{0pt}
\end{tabular}
\vspace*{-0.1cm}
\caption{Comparison of \algname{} across five reasoning selection methods with different selection levels and criteria (Acc.=answer accuracy; PP.=predictive perplexity).}
\label{tab:selection-method-comparison}
\vspace*{-0.1cm}
\end{wraptable}

Table~\ref{tab:selection-method-comparison} summarizes the reasoning quality and distillation performance under five different selection criteria. The results show that our predictive perplexity achieves the highest scores on both reasoning quality metrics. While a high perplexity score is expected, the significantly lower values from other methods indicate their failure to anticipate and guide future reasoning steps effectively. A more direct comparison comes from the performance of student models trained on the resulting reasoning data, where the predictive perplexity-based approach consistently achieves the best results.

Alternative strategies show clear limitations. Random and Max-length Selection introduce noise and fail to ensure reasoning quality. PRM partially filters errors but often removes trajectories that could self-correct into higher-quality reasoning. Binary Judgment provides only discrete labels instead of continuous scores, producing a sparse signal that struggles to capture subtle quality differences.

\subsubsection{Effect of Decoding Strategy}
\label{sec:additional_study_search}

The final component of \algname{} is the decoding strategy, which, rather than relying on local greedy decisions, aims to explore and preserve diverse reasoning paths. We compare \algname{} against two decoding variants, Greedy Decoding and MCTS. For MCTS, we use the same perplexity-based scoring, while utilizing expansion and backpropagation based on upper confidence bound \citep{kocsis2006bandit}. We generate four reasoning trajectories in both variants to match the computational budget.

\begin{wraptable}{r!}{0.55\textwidth}
\vspace*{-0.6cm}
\renewcommand{\arraystretch}{1.1}
\begin{center}
\footnotesize
\begin{tabular}{|L{2cm}|X{0.8cm}|X{0.8cm}|X{0.95cm}|X{0.95cm}|}
\specialrule{\heavyrulewidth}{0pt}{0pt}
{Decoding} &
\multicolumn{2}{c|}{Reasoning Qual.}  &
\multicolumn{2}{c|}{Distillation Perf.} \\ 
% \cmidrule{2-5}
Strategy & 
Acc. & 
PP. & \makecell{\!\!AIME24\!\!} 
& \makecell{\!\!AIME25\!\!} \\ 
%\midrule
%wo. Distillation      & N/A   & N/A   & 71.6 & 53.8 \\ 
\specialrule{\heavyrulewidth}{0pt}{0pt}

Greedy           & 81.6  & \!0.719\!  & 76.7 & 66.5 \\  
MCTS               & 89.6   & \!0.755\!  & 75.8 & 66.3 \\ 
\textbf{Beam Search}          & \textbf{93.1}  & \textbf{\!0.774\!}  & \textbf{79.6} & \textbf{70.2} \\ 
\specialrule{\heavyrulewidth}{0pt}{0pt}
\end{tabular}
\end{center}
\vspace*{-0.3cm}
\caption{Comparison of \algname{} across decoding strategies (Acc.=answer accuracy; PP.=predictive perplexity).
}
\label{table:search_comparision_effectiveness}
\vspace*{-0.3cm}
\end{wraptable}

Table~\ref{table:search_comparision_effectiveness} compares three decoding variants. The results show that beam search delivers the strongest reasoning quality and distillation performance by enabling balanced exploration and collaboration. In contrast, greedy decoding keeps a single hypothesis and enforces locally optimal choices at each step, leading to short-sighted and unstable exploration. MCTS assigns trajectory-level rewards via full rollouts, which makes it less synergistic. Its search biases toward stronger teachers, even when weaker ones are better at specific steps, weakening complementarity (see Appendix~\ref{appendix:decoding_strategy}).

Furthermore, Appendix \ref{appendix:computational_efficiency} analyzes efficiency in terms of wall-clock time against \texttt{Curation} and \texttt{MCTS}. \algname{} runs in roughly half the computation time (49.0\%) of \texttt{MCTS} with negligible meta-prover overhead, and achieves substantially higher reasoning quality than \texttt{Curation} at modest additional cost, demonstrating more effective use of computation.

%, while keeping meta-prover evaluation overhead modest. Compared to Curation, CoRD introduces only a small additional cost, yet delivers significantly higher reasoning quality. Importantly, even when the computation budget of Curation is increased to match that of CoRD by doubling the number of samples, it fails to reach CoRD’s reasoning quality. This indicates that simply increasing sample budgets in post-hoc pipelines leads to substantial computational waste and cannot efficiently produce high-quality reasoning, whereas CoRD achieves superior performance with much lower computational overhead than MCTS and more effective use of computation than Curation.

%Moreover, efficiency analysis shows that beam search is fast in practice, quickly converging to high-quality reasoning trajectories through effective, complementary exploration. It is $1.46\times$ faster than greedy decoding and $2.04\times$ faster than MCTS, while producing high-quality Long-CoT reasoning that are difficult to achieve by merely scaling up curation (See Appendix~\ref{appendix:computational_efficiency} for datailed analysis).

% Notably, \texttt{Curation\textsubscript{$\times 2$}} attains lower predictive perplexity than step-wise decodings without contributing student performance. This indicates that increased post-hoc computation fails to produce higher-quality reasoning, instead causing computational waste.

\subsection{Generalization of CoRD}
\label{sec:generalization}

We apply \algname{} to two additional arithmetic tasks and one open-ended reasoning task to evaluate generalization, as summarized in Table~\ref{tab:student-other-domain}.

\begin{wraptable}{r!}{0.56\textwidth}
\vspace*{-0.7cm}
\renewcommand{\arraystretch}{1.1}
\begin{center}
\setlength{\tabcolsep}{6pt}
\footnotesize
\begin{tabular}{|L{2.9cm}|X{1.2cm}|X{0.8cm}|X{1.3cm}|}
\specialrule{\heavyrulewidth}{0pt}{0pt}
Distillation Pipeline & \!\!MATH500\!\! & \!\!TaTQA\!\! & \!\!\!PubMedQA\!\!\! \\ \specialrule{\heavyrulewidth}{0pt}{0pt}
wo. Distillation        & 92.1 & 87.3 & 86.0 \\ \specialrule{\heavyrulewidth}{0pt}{0pt}

\texttt{Curation-Homo}     & 93.5 & 80.5 & 86.1 \\
\texttt{Integration-Homo}  & 74.1 & 73.3 & 84.0 \\
\textbf{\algname{}\texttt{-Homo}} 
                           & \textbf{93.9} & \textbf{90.0} & \textbf{90.6} \\ \specialrule{\heavyrulewidth}{0pt}{0pt}

\texttt{Curation-Hetero}   & 93.4 & 88.2 & 88.4 \\
\texttt{Integration-Hetero}\!\!\!
                           & 72.3 & 73.1 & 83.0 \\
\textbf{\algname{}\texttt{-Hetero}} 
                           & \textbf{94.8} & \textbf{95.2} & \textbf{91.8} \\ 
\specialrule{\heavyrulewidth}{0pt}{0pt}
\end{tabular}
\end{center}
\vspace*{-0.45cm}
\caption{Distillation performance comparison across three pipelines under two teacher configurations on MATH500, TaTQA, and PubMedQA.}
\label{tab:student-other-domain}
\vspace*{-0.45cm}
\end{wraptable}

\smallskip\smallskip\noindent
{\textbf{Arithmetic Reasoning.}} We test \algname{} and two baselines (trained using R1-Qwen-32B in Table \ref{table:model-size-comparison-final}) on MATH500 \citep{hendrycks2021measuring} and TaTQA \citep{zhu2021tat}. Here, MATH500 shares a similar problem structure with AIME (in-domain), whereas TaTQA requires table-based reading comprehension (out-of-domain). As shown in the 2nd and 3rd columns of Table \ref{tab:student-other-domain}, \algname{} outperforms other methods in Pass@1, indicating that its distilled reasoning transfers robustly beyond AIME.

\smallskip\smallskip\noindent
{\textbf{Open-ended Reasoning.}} To assess \algname{} beyond mathematical tasks, we evaluate it on PubMedQA \citep{jin2019pubmedqa}, an open-domain biomedical QA benchmark with long, free-form answers. Since PubMedQA requires domain-specific, paragraph-level reasoning, we construct a new reasoning-distillation dataset of 456 samples and train a student model (R1-Qwen-32B) accordingly (see Appendix \ref{appendix:detail_pumbedqa} for implementation details for open-ended tasks.). As shown in the 4th column of Table \ref{tab:student-other-domain}, \algname{} achieves the highest Pass@1, demonstrating its effectiveness on open-ended, domain-specific reasoning tasks.

\vspace*{-0.1cm}
\section{Conclusion}
\label{sec:conclusion}
\vspace*{-0.15cm}

We presented \algname{}, which redefines reasoning distillation as a dynamic, step-wise decoding process. By enabling collaborative construction of reasoning trajectories among teacher LRMs, \algname{} produces richer supervision and significantly improves student performance under moderate compute budgets. These results highlight that fine-grained collaboration and progress-aware evaluation are key to efficiently scaling Long-CoT reasoning distillation.

\section*{Limitations}
Our evaluation primarily focused on the monolingual AIME24 and AIME25 benchmarks, and it remains unclear whether the proposed method can generalize to multilingual settings. 
Recent work suggests translating English reasoning traces into other languages to enhance multilingual capabilities, given that large language models (LLMs) are predominantly trained on English corpora \citep{wu2025english}.
We will explore whether our approach can effectively enable cross-lingual transfer of high-quality reasoning in future work.

Additionally, our distillation setup employs only SFT. While our primary focus has been on extracting high-quality reasoning traces, recent studies have explored leveraging preference learning such as direct preference optimization (DPO) to better align models to bridge the disparity between LRMs and suboptimal reasoning patterns like Short-CoT \citep{yang2025thinking}. Extending this line of inquiry, we aim to enhance distillation performance in future work by fostering richer interplay between our high-quality reasoning and complementary preference-aligned datasets.

\section*{Ethical Considerations}

Our work aims to enhance distillation performance through collaborative decoding among LRMs. All training data are generated by publicly available LRMs and do not involve human subjects or sensitive information. Therefore, no additional ethical concerns are raised during the data collection or training phase.

\section*{Scientific Artifacts}

% teacher 3개
% GPT5o
% student 2개 더 

The reasoning generation in our experiments is produced using a total of 4 language models. For open-source models, we utilized publicly available checkpoints from Hugging Face, and for the proprietary model, we accessed them through paid APIs in OpenAI. Detailed model and checkpoint information are provided in Appendix~\ref{appendix:data_generation_param}.

\section*{Acknowledgements}

This research was supported by the National Research Foundation of Korea (NRF) funded by  Ministry of Science and ICT (RS-2022-NR068758), and the "Advanced GPU Utilization Support Program" funded by the Government of the Republic of Korea (Ministry of Science and ICT) (No. 02-26-01-0181). This work was also supported by the National Supercomputing Center with supercomputing resources including technical support (KSC-2025-CRE-0470), and the Korea Basic Science Institute (National research Facilities and Equipment Center) grant funded by the Korea government(MSIT) (No. RS-2026-25492133).

\clearpage
\bibliographystyle{assets/plainnat}
\bibliography{colm2026_conference}

\clearpage
\appendix
\begin{wraptable}{r!}{0.56\textwidth}
\vspace*{-0.6cm}
\renewcommand{\arraystretch}{1.3}
\footnotesize
\begin{center}
\setlength{\tabcolsep}{3pt}
\begin{tabular}{|L{2.8cm}|L{4.4cm}|}
\specialrule{\heavyrulewidth}{0pt}{0pt}
\multicolumn{1}{|l}{Model   Name}        & Checkpoints   \\ \specialrule{\heavyrulewidth}{0pt}{0pt}
\makecell[l]{R1-Qwen-32B\\ \citep{guo2025deepseek}}
& deepseek-ai/DeepSeek-R1-Distill-Qwen-32B  \\
\makecell[l]{QwQ-32B\\ \citep{qwen2.5}} & Qwen/QwQ-32B  \\
\makecell[l]{Phi4-Reasoning-Plus\\ \citep{abdin2025phi}} & microsoft/Phi-4-reasoning-plus     \\
GPT5o-mini             
& gpt5o-mini (OpenAI)            \\
\specialrule{\heavyrulewidth}{0pt}{0pt}
\end{tabular}
\vspace*{-0.25cm}
\caption{Checkpoints of the 4 reasoning generation models. For open-source models, we use publicly available checkpoints from Huggingface, while for proprietary model, we utilize paid API services in OpenAI.}
\label{tab:reasoning-source}
\end{center}
\vspace*{-0.5cm}
\end{wraptable}

\section{Reasoning Generation and Selection Details.}
\label{appendix:data_generation_param}

For reasoning generation in the main experiment, we use the LIMO-v1, LIMO-v2, and S1k-1.1 datasets, which contain 817, 800, and 1000 samples, respectively. We utilize publicly available checkpoints from Hugging Face and paid the API service, as described in Table \ref{tab:reasoning-source}. The user prompt consisted solely of the question text, without any additional context. The system prompt followed the recommended instructions in accordance with each model’s usage guidelines. The prompts used for predictive perplexity evaluation and the Integration baseline are detailed below.

\subsection{Predictive perplexity} 
\label{appendix:perplexity}
Predictive perplexity is computed as described in Section \ref{sec:perplexity}. We insert the partial reasoning and ground-truth answer into the prompt to calculate the predictive perplexity below:
 
\begin{tcolorbox}[colback=white!90!, colframe=black!100, title=Predictive Perplexity Calculation Template]
\textcolor{blue}{$\tau_{<t}\oplus s_t^k$} \texttt{</think>} The final answer is~\textcolor{blue}{\{GT\}}.
\end{tcolorbox}

\subsection{Integration Prompt}
\label{appendix:integration}

Table~\ref{tab:integrated-reasoning-prompt} presents the prompt used in the Integration baseline. This prompt is designed to integrate individual reasoning outputs from multiple LRMs in a manner consistent with the Long-CoT framework. It guides the integrator to merge reasoning steps into a coherent reasoning trace that preserves the characteristics emphasized in LRMs.

\begin{wraptable}{r!}{0.56\textwidth}
\vspace*{-0.6cm}
\begin{center}
\renewcommand{\arraystretch}{1.1}
\footnotesize
\begin{tabular}{|L{2.5cm}|X{0.7cm}|X{0.6cm}|X{0.8cm}|X{0.8cm}|}
\specialrule{\heavyrulewidth}{0pt}{0pt}
{\!\!Meta-prover} &
\multicolumn{2}{c|}{\!\!Reasoning Qual.\!\!} &
\multicolumn{2}{c|}{\!\!Distillation Perf.\!\!} \\ 
\!\!Models &
Acc. &
PP.\!\!\!\! &
\makecell{\!\!\!\!AIME24\!\!\!\!} &
\makecell{\!\!\!\!AIME25\!\!\!\!} \\ 
\specialrule{\heavyrulewidth}{0pt}{0pt}

\!\!QwQ-32B (Strong)   & \!93.1\! & 0.774\!\!\!\! & \!79.6\! & \!70.2\! \\ 
\!\!Phi-4 (Moderate)  & \!89.2\! & 0.749\!\!\!\! & \!75.9\! & \!64.4\! \\ 
\!\!R1-Qwen (Weak)    & \!80.5\! & 0.641\!\!\!\! & \!68.5\! & \!53.2\! \\ 

\specialrule{\heavyrulewidth}{0pt}{0pt}
\end{tabular}
\end{center}
\vspace*{-0.1cm}
\caption{Effect of meta-prover choice on reasoning quality and distillation performance. Strength is assigned based on Pass@1 performance on AIME24 and AIME25 (Acc.=answer accuracy; PP.=predictive perplexity).}
\label{tab:meta-prover-comparison}
\vspace*{-0.1cm}
\end{wraptable}

\section{Results with Other Meta-provers}
\label{appendix:other_meta_prover}

For step-level guidance via predictive perplexity, we adopt the strongest teacher model as the meta-prover. Unlike approaches that rely on a trained external reward model, this design introduces no additional training or deployment dependency, as the teacher model is already available during distillation. Nevertheless, a natural question arises as to whether a stronger meta-prover is always the most appropriate source of guidance. In particular, weaker models may provide more compatible supervision, as different architectures can be specialized for different types of tasks. To examine this possibility, we evaluate alternative meta-provers with varying strengths and analyze their impact on both reasoning quality and distillation performance, as summarized in Table~\ref{tab:meta-prover-comparison}.

This results shows that weaker meta-provers reduce reasoning quality and distillation performance, which is an expected outcome in knowledge distillation. This highlight the importance of carefully selecting the meta-prover from the teacher pool, as the choice of meta-prover can affect both reasoning quality and distillation performance.

\section{Training Details}
\label{appendix:train_and_inference_param}

\begin{wraptable}{r!}{0.56\textwidth}
\vspace*{-0.7cm}
\footnotesize
\begin{center}
\setlength{\tabcolsep}{20pt}
\renewcommand{\arraystretch}{1.1}
\begin{tabular}{|l|l|}
\specialrule{\heavyrulewidth}{0pt}{0pt}
\multicolumn{1}{|l|}{Parameter}           & Value                         \\ \specialrule{\heavyrulewidth}{0pt}{0pt}
Batch size                              & 8                             \\
Epochs                                  & 5                             \\
Learning rate                           & 5.0e-6                          \\
Max sequence length                     & 20480                          \\
LR scheduler type                       & cosine                        \\ \specialrule{\heavyrulewidth}{0pt}{0pt}
\end{tabular}
\vspace{-0.2cm}
\caption{Hyperparameters of the training configuration.}
\label{table:hyperparameters}
\end{center}
\vspace{-1cm}
\end{wraptable}

We fine-tune the student model using supervised fine-tuning (SFT) with DeepSpeed (Stage-3) \citep{rasley2020deepspeed} on 8 × NVIDIA H100 GPUs. Table~\ref{table:hyperparameters} summarizes the training configurations for SFT. During training, reasoning trajectories are enclosed within \texttt{<think>} tags.

\section{Results with Other Homogeneous Teacher Model Setups}
\label{appendix:results1_remaing_results}

\begin{table*}[t!]
\begin{center}
\renewcommand{\arraystretch}{1.1}
\footnotesize
\begin{tabular}{|L{2cm}|L{2.7cm}|X{1.4cm}|X{1.4cm}|X{1.8cm}|X{1.8cm}|}
\specialrule{\heavyrulewidth}{0pt}{0pt}
\multirow{2}{*}{Framework} & 
\multirow{2}{*}{Teacher Model} & 
\multirow{2}{*}{\makecell{Answer \\ Accuracy}} & 
\multirow{2}{*}{\makecell{Predictive \\ Perplexity}} & 
\multicolumn{2}{c|}{Distillation Performance} \\ 
\cline{5-6}
& & & & \makecell{AIME24} & \makecell{AIME25} \\ 
\specialrule{\heavyrulewidth}{0pt}{0pt}
\multicolumn{2}{|c|}{Before Training} & N/A & N/A & 71.6 & 53.8 \\ 
\specialrule{\lightrulewidth}{0pt}{0pt}
\multirow{3}{*}{\texttt{Curation}} 
& QwQ-32B              & 77.4 & 0.664 & 74.2 & 62.7 \\ 
& R1-Qwen-32B          & 49.6 & 0.415 & 62.9 & 47.9 \\ 
& Phi4-Reasoning-Plus  & 67.8 & 0.527 & 71.3 & 60.8 \\ 
\specialrule{\lightrulewidth}{0pt}{0pt}
\multirow{3}{*}{\texttt{Integration}} 
& QwQ-32B              & 88.6 & 0.215 & 11.9 & 6.9 \\ 
& R1-Qwen-32B          & 70.1 & 0.319 & 8.5  & 5.8 \\ 
& Phi4-Reasoning-Plus  & 64.2 & 0.310 & 7.4  & 5.6 \\ 
\specialrule{\lightrulewidth}{0pt}{0pt}
\multirow{3}{*}{\algname{}} 
& QwQ                  & 90.0 & 0.726 & 75.8 & 64.4 \\ 
& R1-Qwen              & 73.2 & 0.573 & 69.8 & 56.0 \\ 
& Phi4-Reasoning-Plus  & 84.0 & 0.628 & 72.5 & 63.9 \\ 
\specialrule{\lightrulewidth}{0pt}{0pt}
\end{tabular}
\end{center}
\vspace*{-0.4cm}
\caption{Reasoning data quality and distillation performance in homogeneous settings.}
\label{table:homogeneous_settung}
\vspace*{-0.4cm}
\end{table*}

Table~\ref{table:homogeneous_settung} presents the results in homogeneous (single-teacher) settings. \algname{} demonstrates that even in single-teacher settings, collective step-wise decoding consistently improves overall data quality across all teacher models, surpassing \texttt{Curation} and \texttt{Integration} in every case. This confirms that its advantages arise from organized reasoning rather than stochastic diversity under the compute budget.

\section{Results with R1-Llama-8B}
\label{appendix:results1_other_model}

\begin{wraptable}{r!}{0.56\textwidth}
\vspace*{-0.6cm}
\begin{center}
\setlength{\tabcolsep}{3pt}
\renewcommand{\arraystretch}{1.1}
\footnotesize
\begin{tabular}{|L{3cm}|X{1.8cm}|X{1.8cm}|}
\specialrule{\heavyrulewidth}{0pt}{0pt}
Distillation Pipeline & 
AIME24 & 
AIME25 \\ 
\specialrule{\heavyrulewidth}{0pt}{0pt}
wo. Distillation        & 46.5 & 31.8\\ 
\specialrule{\heavyrulewidth}{0pt}{0pt}
\texttt{Curation-Homo}          & 48.5 & 33.7 \\ 
\texttt{Integration-Homo}       & 1.4 & 2.0 \\ 
\textbf{\algname{}\texttt{-Homo}}        & \textbf{50.4} & \textbf{37.7} \\ \specialrule{\heavyrulewidth}{0pt}{0pt}
\texttt{Curation-Hetero}        & 41.3 & 30.8 \\ 
\texttt{Integration-Hetero}     & 1.0 & 0.2 \\ 
\textbf{\algname{}\texttt{-Homo}}             & \textbf{54.0} & \textbf{39.8} \\ 
\specialrule{\heavyrulewidth}{0pt}{0pt}
\end{tabular}
\end{center}
\vspace*{-0.2cm}
\caption{Distillation performance comparison of R1-Llama-8B model across six frameworks.}
\label{table:model-size-comparison-7b}
\vspace*{-0.4cm}
\end{wraptable}

We conduct an additional experiment to examine whether the benefits from \algname{} generalize to different LRM families. Specifically, we evaluate DeepSeek-R1-Distill-Llama-8B, whose architecture and pretraining pipeline differ from the Qwen-based teachers (QwQ and R1-Qwen) used in the main experiments. As shown in Table~\ref{table:model-size-comparison-7b}, \algname{} consistently outperforms all baseline frameworks, confirming that the overall trends remain consistent and that it provides strong, stable distillation signals even for models outside the Qwen family.

\section{Post-hoc Integration with Stronger Integrator}
\label{appendix:result_strong_integration}

For the \texttt{Integration} baseline, we select GPT5o-mini as the integrator due to its architectural difference from the teacher model pool, which helps avoid bias, as well as its comparable performance to the teachers, thereby preventing confounding effects from a strong integrator. 
Despite this choice, and despite all pipelines using the same student model and scoring procedure, the student model trained under the \texttt{Integration} baseline exhibits a significant performance degradation prior to training. To further investigate whether this degradation is related to integrator capacity, we replace GPT-5o-mini with a stronger integrator, DeepSeek-V3.2-Exp, within the heterogeneous teacher integration setting.

\begin{wraptable}{r!}{0.56\textwidth}
\vspace*{-0.6cm}
\begin{center}
\renewcommand{\arraystretch}{1.1}
\footnotesize
\begin{tabular}{|L{2.5cm}|X{0.7cm}|X{0.6cm}|X{0.8cm}|X{0.8cm}|}
\specialrule{\heavyrulewidth}{0pt}{0pt}
{\!\!Integrator} &
\multicolumn{2}{c|}{\!\!Reasoning Qual.\!\!} &
\multicolumn{2}{c|}{\!\!Distillation Perf.\!\!} \\ 
\!\!Models &
Acc. &
PP.\!\!\!\! &
\makecell{\!\!\!\!AIME24\!\!\!\!} &
\makecell{\!\!\!\!AIME25\!\!\!\!} \\ 
\specialrule{\heavyrulewidth}{0pt}{0pt}

\!\!GPT-5o-mini & \!91.2\! & \!0.223\!\!\!\! & \!12.7\! & \!9.0\! \\
\!\!DeepSeek-V3.2-Exp & \!96.2\! & \!0.199\!\!\!\! & \!17.3\! & \!12.9\! \\

\specialrule{\heavyrulewidth}{0pt}{0pt}
\end{tabular}
\end{center}
\vspace*{-0.4cm}
\caption{Comparison of reasoning quality and distillation performance across two integrators for the \texttt{Integration} baseline in the heterogeneous teacher configuration.}
\label{table:integrator-comparison}
\vspace*{-0.5cm}
\end{wraptable}

Table~\ref{table:integrator-comparison} summarizes the performance of the \texttt{Integration} baseline across different integrators in the heterogeneous teacher configuration. Although a stronger integrator yields modest improvements, it still fails to reconstruct coherent Long-CoT structures, suggesting that the limitation does not primarily stem from integrator weakness or implementation issues, but rather reflects a fundamental challenge in post-hoc integration for current LLMs. In particular, processing extremely long contexts remains difficult due to lost-in-the-middle~\citep{liu2024lost} and needle-in-a-haystack~\citep{kuratov2024babilong} effects. In the \texttt{Integration} baseline, the integrator must aggregate nearly 30K tokens of teacher Long-CoT reasoning into a coherent trajectory exceeding 4K tokens, which often leads to a collapse into short and shallow Short-CoT outputs. Post-hoc integration at this scale remains unreliable with existing methods, motivating the use of \algname{}’s dynamic, step-wise synthesis.

\section{Additional Experiment Details}
\label{sec:additional_detail}

\begin{figure*}[t]
\begin{center}
\includegraphics[width=13cm]{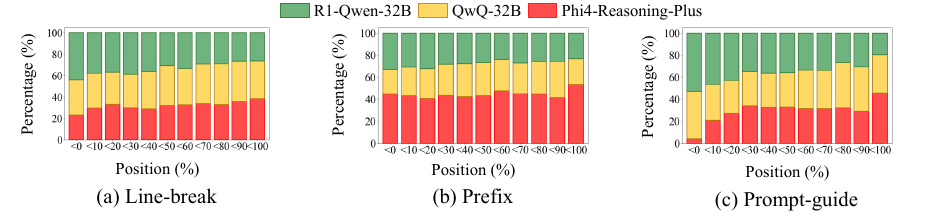}
% \vspace*{-0.5cm} % 하단 여백 추가
\end{center}
\vspace*{-0.6cm}
\caption{Hit rates of three LRMs during expansion across three step units based on step locations that reflect their relative position ratios within the entire reasoning.}
\label{fig:results3_hitrate}
\vspace*{-0.4cm}
\end{figure*}

\subsection{Analysis of Reasoning Dynamics Across Step Segmentations}
\label{app:step_unit_example}

We analyze how different step segmentation schemes affect reasoning structure and multi-teacher collaboration.

\smallskip\smallskip\noindent 
\textbf{Step unit Configuration.} 
For the step unit selection, the line-break unit simply matches the word {\textbackslash n\textbackslash n} as a boundary, whereas the prefix-based unit requires matching prefixes corresponding to reasoning patterns in Long-CoT reasoning. Following \citep{ye2025limo} and \citep{qu2025optimizing}, we selected appropriate prefix terms as follows:

\begin{itemize}
    \item Self-Verification: "let me check", "let me verify", "double-check", "going back to", "wait"
    \item Multi-method Validation: "alternatively", "another way", "let's try a different approach", "using another method", "we can also verify"
    \item Self-Correction: "this is wrong", "the mistake was", "that's impossible", "this contradicts", "the error is"
\end{itemize}

\smallskip\smallskip\noindent 
\textbf{Comparison of Collaboration Dynamics.}
We further analyze collaboration dynamics across different step units in Figure~\ref{fig:results3_hitrate}, which reports each LRM’s hit rate during reasoning expansion.
We note that early reasoning stages typically involve problem formulation and constraint analysis \citep{he2025can}. Prompt-guided step units align with these semantic phases, enabling heterogeneous LRMs to collaborate at stages best suited to their strengths. 
In this setting, QwQ and R1-Qwen dominate the early steps, while Phi4-Reasoning-Plus contributes more in later stages that require comprehending prior steps for conclusion. 
In contrast, prefix step units are dominated by a few models, with R1-Qwen selected only about 20--25\%, and line-break ones exhibits a similar trend. While line-break step units encourage some stylistic sharing across models, they remain limited in fostering genuine semantic collaboration.
% 참고 다른 데이터셋 (Limov2)에서도 비슷한 추세 보이는거 보니 global한듯

\subsection{Binary Judgement Prompt Details}
\label{sec:results_step_selection}

For binary judgment, we adopt the meta-prover prompt from \citet{qu2025optimizing}, where the judge completes the current prefix into a final answer and assigns a binary correctness score; results are averaged over 10 independent runs to reduce variance. During rollout, we append the phrase "The final answer is" at the end of the reasoning process to encourage the model to quickly and explicitly produce the final answer without additional unnecessary reasoning steps. The prompt is shown below:

\begin{tcolorbox}[colback=white!90!, colframe=black!100, title=Binary Judgment Prompt]
\textcolor{blue}{$\tau_{<t}\oplus s_t^k$} Time is up. Given the time I've spent and the approaches I've tried, I should stop thinking and formulate a final answer based on what I already have.\texttt{</think>} The final answer is:
\end{tcolorbox}

\subsection{Analysis of Reasoning Dynamics across Decoding Strategies}
\label{appendix:decoding_strategy}

\begin{wrapfigure}{r}{0.55\textwidth}
\vspace*{-0.7cm}
\begin{center}
\includegraphics[width=7.7cm]{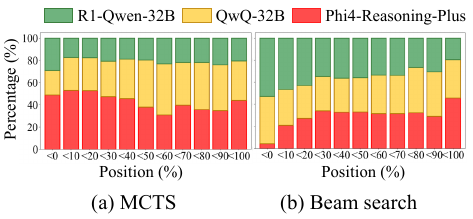}
% \vspace*{-0.5cm} % 하단 여백 추가
\end{center}
\vspace*{-0.3cm}
\caption{Comparison of hit rates during expansion between beam search and MCTS. The step locations reflect their relative position ratios within the entire reasoning.}
\label{fig:hit_rate_mcts}
\vspace*{-0.3cm}
\end{wrapfigure}

We analyze the resulting collaboration dynamics among multiple reasoning trajectories induced by this strategy.
Figure~\ref{fig:hit_rate_mcts} compares the hit-rate distributions of different teachers across reasoning positions under MCTS and beam search, revealing how the decoding strategy alters collaborative dynamics. 
In MCTS, trajectory-level rewards cause the search to converge toward globally stronger teachers, reducing exploration of weaker yet occasionally effective ones. Consequently, complementary reasoning diminishes. 
In contrast, beam search maintains a more balanced mixture of teacher contributions throughout, preserving complementary reasoning behaviors. Interestingly, during the early reasoning steps, beam search leverages the local strengths of weaker teachers such as R1-Qwen-32B, which often provide useful intermediate reasoning cues before stronger teachers dominate in later stages. 
Overall, these results show that MCTS biases collaboration toward high-performing teachers, whereas beam search sustains broader cooperation across reasoning steps.

%%%% 임시 공간

\subsection{Computational Efficiency Analysis}
\label{appendix:computational_efficiency}

For a more comprehensive assessment, we analyze computational efficiency to characterize how different distillation pipelines trade off compute against reasoning quality and distillation performance.
% Specifically, we (i) compare the empirical wall-clock cost per question of \texttt{Curation}, \texttt{MCTS}, and \algname{}, and (ii) perform a compute-matched study to test whether simply increasing post-hoc compute can close the performance gap to \algname{}.

\smallskip\smallskip\noindent
\subsubsection{Wall-clock Time Analysis}

\begin{wraptable}{r!}{0.56\textwidth}
\vspace*{-0.6cm}
\begin{center}
\renewcommand{\arraystretch}{1.1}
\footnotesize
\begin{tabular}{|L{1.5cm}|X{1.5cm}|X{1.5cm}|X{1.5cm}|}
\specialrule{\heavyrulewidth}{0pt}{0pt}
{Distillation Pipeline} &
\makecell{\!\!Step\\Generation\\(A)\!\!} &
\makecell{\!\!Meta-prover\\Evaluation\\(B)\!\!} &
\makecell{\!\!Total\\Generation\\(A+B)\!\!} \\
\specialrule{\heavyrulewidth}{0pt}{0pt}

\!\!\texttt{Curation}\!\! & \!\!167.1\!\! & \!\!1.2\!\!  & \!\!168.3\!\! \\ 
\!\!\texttt{MCTS}\!\!    & \!\!567.7\!\! & \!\!21.5\!\! & \!\!589.2\!\! \\
\!\!\algname{}\!\!\!           & \!\!277.3\!\! & \!\!11.4\!\! & \!\!288.7\!\! \\
\specialrule{\heavyrulewidth}{0pt}{0pt}
\end{tabular}
\end{center}
\vspace*{-0.4cm}
\caption{Computational efficiency comparison of three distillation pipelines. Time reported as the average wall-clock time (in seconds) per question measured on NVIDIA H200$\times4$ GPUs. Breakdown shown for step generation (A) and meta-prover evaluation (B).}
\label{table:search_comparision_time_only}
\vspace*{-0.2cm}
\end{wraptable}

We evaluate the empirical computational efficiency of three distillation pipelines, \texttt{Curation}, \texttt{MCTS}, and \algname{}. As described in Section \ref{sec:computation}, these pipelines primarily differ in how they allocate computation between trajectory generation and verification. \texttt{Curation} generates complete trajectories in a single pass and applies only lightweight post-hoc scoring, yielding a moderate generation cost with negligible verification overhead. In contrast, \texttt{MCTS} repeatedly performs full rollouts and evaluates candidate trajectories, which substantially increases both generation and verification costs. \algname{} generates a comparable number of trajectories to \texttt{Curation} but employs step-wise decoding with cached key-value states and lightweight meta-prover scoring for a small ground-truth answer token, thereby maintaining moderate costs for both generation and verification.

Table \ref{table:search_comparision_time_only} shows the empirical wall-clock computational cost per question under the heterogeneous teacher configuration, measured on NVIDIA H200$\times4$ GPUs. Relative to \texttt{Curation}, \algname{} adds only modest computational cost, as it avoids full rollouts by advancing at the step-level and invokes the meta-prover far less frequently than \texttt{MCTS}. While \algname{} is slightly more expensive than \texttt{Curation}, the additional cost incurred by the meta-prover is small, and the substantial improvement in reasoning quality makes the extra cost reasonable.

\begin{wraptable}{r!}{0.56\textwidth}
\vspace*{-0.6cm}
\begin{center}
\renewcommand{\arraystretch}{1.1}
\footnotesize
\begin{tabular}{|L{1.5cm}|X{0.7cm}|X{0.6cm}|X{0.8cm}|X{0.8cm}|X{0.6cm}|}
\specialrule{\heavyrulewidth}{0pt}{0pt}
{\!\!Decoding\!\!} &
\multicolumn{2}{c|}{\!\!Reasoning Qual.\!\!}  &
\multicolumn{2}{c|}{\!\!Distillation Perf.\!\!} &
\!\!Time\!\! \\ 
\!\!Strategy\!\! & 
\!\!Acc.\!\! & 
PP.\!\!\!\! & 
\makecell{\!\!\!\!AIME24\!\!\!\!} &
\makecell{\!\!\!\!AIME25\!\!\!\!} &  
\!\!Sec\!\! \\ 
\specialrule{\heavyrulewidth}{0pt}{0pt}
\!\!\texttt{Curation}\!\!   & 84.8 & 0.652\!\!\!\! & \!75.0\! & \!62.1\! & \!\!168.3\!\! \\ 
\!\!\texttt{Curation\textsubscript{$\times 2$}}\!\!      & 90.3 & 0.712\!\!\!\! & \!74.6\! & \!63.8\! & \!\!336.6\!\! \\  
%\!\!\texttt{MCTS}\!\!          & 89.6 & 0.755\!\!\!\! & \!75.8\! & \!66.3\! & \!\!589.2\!\! \\ 
\textbf{\!\!\algname{}\!\!\!} 
                      & \textbf{93.1} 
                      & \textbf{0.774\!\!\!\!} 
                      & \textbf{\!79.6\!} 
                      & \textbf{\!70.2\!} 
                      & \textbf{\!\!288.7\!\!} \\ 
\specialrule{\heavyrulewidth}{0pt}{0pt}
\end{tabular}
\end{center}
\vspace*{-0.4cm}
\caption{Reasoning quality and distillation performance comparison across four methods (Acc.=answer accuracy; PP.=predictive perplexity). \texttt{Curation $\times$2} denotes a variant of \texttt{Curation} whose computation budget is increased to match that of \algname{}.}
\label{table:search_comparision}
\vspace*{-0.4cm}
\end{wraptable}

\subsubsection{Equal Computation Budget Analysis} 
We further evaluate whether increasing the compute budget allows \texttt{Curation} to match the reasoning quality of \algname{}. We increase the computation budget for \texttt{Curation} by doubling the number of completions from four to eight, thereby doubling its total generation cost from 168.3s to match that of \algname{} (288.7s), as shown in Table \ref{table:search_comparision_time_only}. Thus, in this increased setup (\texttt{Curation x2)}), the best reasoning trajectory is selected among the eight candidates based on predictive perplexity and then use it to train the student model (R1-Qwen-32B).

Table \ref{table:search_comparision} shows the reasoning quality and distillation performance comparison under the same heterogeneous teacher configuration. \algname{} achieves the best balance between efficiency and performance by allocating compute within a single run, effectively balancing exploration and exploitation.
However, even when we match the compute of \texttt{Curation} to that of \algname{}, the predictive perplexity of \texttt{Curation x2} remains below that of \texttt{MCTS} and \algname{}, and the corresponding student performance does not bring improvement. This indicates that post-hoc pipelines cannot efficiently yield higher quality reasoning even with increased computation, rather leading to many discarded trajectories and substantial computational waste.

\section{Additional Experimental Details for PubMedQA}
\label{appendix:detail_pumbedqa}

Unlike the mathematical domain in LIMO-v1, PubMedQA requires domain-specific, paragraph-level reasoning grounded in scientific evidence and long, free-form conclusions. Because this task demands qualitatively different reasoning capabilities, we construct a new distillation dataset tailored to PubMedQA and train a student model. We conduct the same comparison across \texttt{Curation}, \texttt{Integration}, and \algname{} under an evaluation setup designed for open-ended answers.

\smallskip\smallskip\noindent
\textbf{Dataset.} 
To match LIMO-v1’s question configuration for effective Long-CoT distillation, we follow LIMO-v1 and retain only difficult and complex questions, where difficulty is operationalized as a low success rate and complexity is proxied by reasoning length. Specifically, for each question we sample three complete reasoning trajectories at temperature 0.6 from either Llama-3.1-8B-Instruct or Qwen2.5-7B-Instruct, without explicitly constraining the generation length. We keep questions where all three samples are incorrect and the reasoning length exceeds 1K tokens (counted using R1-Qwen-32B) for complexity. This filtering yields 456 questions from the initial 213.3K instances.

\smallskip\smallskip\noindent
\textbf{Reasoning Data Distillation.} We distill reasoning data using two post-hoc baselines (\texttt{Curation} and \texttt{Integration}) and our \algname{}. For a controlled comparison, we keep the distillation procedure and all hyperparameters identical across methods (\textit{e.g.}, the teacher model pool, sampling settings, and student training configurations), and vary only the generation and meta-prover prompts. We provide a golden grounded paragraph in the reasoning generation prompt to factor out retrieval ability and isolate the quality of the reasoning itself. In the meta-prover prompt, we include the dataset’s reference long-form answer as the target for computing predictive perplexity. We train R1-Qwen-32B on each constructed dataset via supervised fine-tuning.

\smallskip\smallskip\noindent
\textbf{Evaluation.} We use an LLM-as-a-judge with Qwen3-32B to assess answer accuracy and student performance, since exact matching cannot capture diverse valid linguistic formulations in open-ended tasks. We adopt the LLM-as-a-judge prompt from prior work~\citep{choi2025word2passage,song2025ext2gen}, which assigns a binary label for response appropriateness. The prompt is shown below:

\begin{tcolorbox}[colback=white!90!, colframe=black!100, title=Evaluation Prompt for PubmedQA]

Your task is to evaluate the correctness of the predicted answer based on the true answer. \\

Instructions:

1. Read the QUERY and then compare the ANSWER and the Predicted ANSWER. 

2. Check if the Predicted Answer includes the core content of the True Answer (True/False in text).

3. If the Predicted Answer is correct, return "True". If it is incorrect, return "False". 

QUERY: 
\textcolor{blue}{\{Question\}}

TRUE ANSWER: 
\textcolor{blue}{\{Reference Answer\}}

Predicted ANSWER: 
\textcolor{blue}{\{Model Answer\}
}

Output Format: 
\{"correctness": "True or False"\}

Output (Only JSON):

\end{tcolorbox}

\begin{wraptable}{r!}{0.58\textwidth}
\vspace*{-0.6cm}
\begin{center}
\renewcommand{\arraystretch}{1.1}
\footnotesize
\begin{tabular}{|L{2.8cm}|X{0.8cm}|X{0.4cm}|X{1.8cm}|}
\specialrule{\heavyrulewidth}{0pt}{0pt}
\!\!Selection &
\multicolumn{2}{c|}{Reasoning Qual.} &
\!\!\!Distillation Perf.\!\!\! \\ 
\!\!Method &
Acc. &
PP.\!\!\!\!\!\!\!\! &
\makecell{PubMedQA} \\ 
\specialrule{\heavyrulewidth}{0pt}{0pt}

\!\!wo. Distillation
& N/A
& N/A\!\!\!\!\!\!\!\!
& 86.0 \\ 
\specialrule{\heavyrulewidth}{0pt}{0pt}

\!\!\texttt{Curation-Homo}\!\!       & 62.6 & 0.180\!\!\!\!\!\!\!\! & 86.1 \\ 
\!\!\texttt{Integration-Homo}\!\!   & 65.4 & 0.216\!\!\!\!\!\!\!\! & 84.0 \\ 
\!\!\texttt{\algname{}-Homo}\!\!          & 70.3 & 0.284\!\!\!\!\!\!\!\! & 90.6 \\ 
\specialrule{\heavyrulewidth}{0pt}{0pt}
\!\!\texttt{Curation-Hetero}\!\!    & 71.4 & 0.243\!\!\!\!\!\!\!\! & 88.4 \\ 
\!\!\texttt{Integration-Hetero}\!\!\!\!\!\! & 65.6 & 0.215\!\!\!\!\!\!\!\! & 83.0 \\ 
\!\!\texttt{\algname{}-Hetero}\!\!
                    & \textbf{75.8} & \textbf{0.339}\!\!\!\!\!\!\!\! & \textbf{91.8} \\ 
\specialrule{\heavyrulewidth}{0pt}{0pt}
\end{tabular}
\end{center}
\vspace*{-0.4cm}
\caption{Quality of the generated reasoning and distillation performance comparison across three distillation pipelines under two teacher configurations.}
\label{table:pipeline-comparison-qa}
\vspace*{-0.5cm}
\end{wraptable}

\smallskip\smallskip\noindent 
\textbf{Reasoning Quality Comparison.} 
Table~\ref{table:pipeline-comparison-qa} presents reasoning quality and distillation performance (R1-Qwen-32B) across three distillation pipelines under two teacher configurations. \algname{} consistently produces higher-quality reasoning traces and achieves stronger distillation performance than the baselines, and the same relationship between reasoning quality and performance observed in math domains also holds for open-domain task.

%%%%% 임시공간

% \subsection{Effect of MCTS}
% \label{appendix:additional_decording}

\begin{table*}[t]
\centering
\begin{tabular}{p{0.97\linewidth}}
\toprule
\textbf{Instruction} \\ \midrule

You are tasked with analyzing multiple reasoning solutions and integrating them into a single, structured JSON output. \\

1. Integrate All Reasoning \\
- The reasoning steps are provided inside XML tags such as: \\
<reasoning\_step\_1> ... </reasoning\_step\_1> \\
<reasoning\_step\_2> ... </reasoning\_step\_2> \\
- Merge the content inside all these XML tags into one unified reasoning flow. \\
- Combine them carefully while maintaining logical flow and context. \\

\vspace{0.2cm}
2. Assign IDs \\
- Each sub-thinking process should have its own unique ID. \\
- Use a hierarchy such as: \\
"integrated\_step1", "integrated\_step2" for overall stages of integrated reasoning. \\
"answer\_part" for the final answer section and use \textbackslash boxed\{\} format for the final answer. \\

\vspace{0.2cm}
3. Categorize Reasoning Patterns \\
Categorize the reasoning according to its type to ensure effective integration: \\
- Progressive Reasoning: Logical, forward-moving step-by-step problem solving. \\
  Indicators: “Let's solve”, “First”, “Next”, “Then”, “Therefore”, “We need to”, “Given that”. \\
- Verification: Returning to check previous steps for accuracy. \\
  Indicators: “Wait”, “Let me check”, “Let me verify”, “Double-check”, “Going back to”. \\
- Multi-method Validation: Using different methods or perspectives to confirm a conclusion. \\
  Indicators: “Alternatively”, “Another way”, “Let's try a different approach”, “Using another method”. \\
- Error Correction Pattern: Identifying and fixing mistakes in reasoning. \\
  Indicators: “This is wrong”, “The mistake was”, “This can't be right”, “The error is”, “This contradicts”. \\

\vspace{0.2cm}
4. Return Your Integration in JSON Format \\
Provide your integrated reasoning in the following JSON structure: \\[2pt]

\texttt{\{} \\
\hspace{0.5cm}"integrated\_step1": \{"content": "\#\#\# Step 1. <integrated reasoning text>", "category": "<reasoning pattern>"\}, \\
\hspace{0.5cm}"integrated\_step2": \{"content": "\#\#\# Step 2. <integrated reasoning text>", "category": "<reasoning pattern>"\}, \\
\hspace{0.5cm}... \\
\hspace{0.5cm}"integrated\_stepN": \{"content": "\#\#\# Step N. <integrated reasoning text>", "category": "<reasoning pattern>"\}, \\
\hspace{0.5cm}"answer\_part": ["<final answer in boxed format>"] \\
\texttt{\}} \\[2pt]

\midrule
\textbf{Target Input Example} \\
\midrule
\textbf{Question:} 
{\color{blue}\{Question\}} \\[2pt]

\textbf{Reasoning Steps:} 
{\color{blue}\{Reasoning Steps\}} \\[2pt]

\bottomrule
\end{tabular}
\caption{Prompt used to integrate reasoning across LRMs.}
\label{tab:integrated-reasoning-prompt}
\end{table*}

\end{document}